# Can Foundation Models Reliably Identify Spatial Hazards? A Case Study on Curb Segmentation

Diwei Sheng[a], Giles Hamilton-Fletcher[b,d], Mahya Beheshti[b,g], Chen Feng*[a,g] and John-Ross Rizzo*[b,c,d,e,f,g,h]

a) Department of Computer Science and Engineering, NYU Tandon School of Engineering, Brooklyn, NY 11201, USA
b) Department of Rehabilitation Medicine, NYU Grossman School of Medicine, New York, NY 10016, USA
c) Department of Rehabilitation Medicine, NYU Langone Health, New York, USA
d) Department of Ophthalmology, NYU Langone Health, New York, USA
e) Department of Neurology, NYU Langone Health, New York, USA
f) Department of Biomedical Engineering, NYU Tandon School of Engineering, New York, USA
g) Department of Mechanical & Aerospace Eng., NYU Tandon School of Engineering, New York, USA
h) Institute for Excellence in Health Equity, New York University Grossman School of Medicine, New York, USA

*Authors to whom correspondence should be addressed

# Can Foundation Models Reliably Identify Spatial Hazards? A Case Study on Curb Segmentation

Abstract— Curbs serve as vital borders that delineate safe pedestrian zones from potential vehicular traffic hazards. Curbs also represent a primary spatial hazard during dynamic navigation with significant stumbling potential. Such vulnerabilities are particularly exacerbated for persons with blindness and low vision (PBLV). Accurate visual-based discrimination of curbs is paramount for assistive technologies that aid PBLV with safe navigation in urban environments. Herein, we investigate the efficacy of curb segmentation for foundation models. We introduce the largest curb segmentation dataset to-date to benchmark leading foundation models. Our results show that state-of-the-art foundation models face significant challenges in curb segmentation. This is due to their high false-positive rates (up to 95%) with poor performance distinguishing curbs from curb-like objects or non-curb areas, such as sidewalks. In addition, the best-performing model averaged a 3.70-second inference time, underscoring problems in providing real-time assistance. In response, we propose solutions including filtered bounding box selections to achieve more accurate curb segmentation. Overall, despite the immediate flexibility of foundation models, their application for practical assistive technology applications still requires refinement. This research highlights the critical need for specialized datasets and tailored model training to address navigation challenges for PBLV and underscores implicit weaknesses in foundation models.

**Introduction**

Curbs are crucial landmarks, delineating the boundary between pedestrian areas and vehicular lanes, and serve as navigational cues for persons navigating the built environment. These spatial cues are critical for persons who are blind or have low vision (PBLV) (Koutsoklenis and Papadopoulos, 2014; Ståhl et al., 2010). These tactile markers serve as indicators for safe walking spaces and signal the transition to areas where vehicular traffic poses a greater risk (Sauerburger, 2005). By consistently defining the perimeter of pedestrian zones, curbs provide a tactile guide, enhancing the mobility and safety of those with visual impairments (Bentzen et al., 2017; Norgate, 2012; Childs et al., 2010). This distinction is crucial in urban and suburban environments where navigating between sidewalks and streets is a constant challenge.

While curbs are important for highlighting changes in elevation and specific zoning, it is crucial to emphasize that they are also significant spatial hazards, inducing tripping during navigation for PBLV (Zhao et al., 2018). Curbs are often highly variable in height, width, and spatial geometries, leading to higher trips and falls rates, if not approached with caution and anticipation (Ayres and Kelkar, 2006). Thus, curbs pose a notable risk for accidents due to their stumble potential.

To effectively assist PBLV, computer-vision models must accurately segment and not just detect curbs. This is because establishing the curbs' visual position, orientation and shape is crucial for the user to accurately understand and appropriately approach the curb. By contrast, detecting curbs with bounding boxes do not provide this information and in many scenarios will include a substantial proportion of 'non-curb' area within the detected area, making this feedback difficult to parse and ultimately leverage from the user's perspective. However, object detection can still serve as a preliminary step to reach more pixel-accurate curb segmentation (Lempitsky, 2009; Dai et al., 2015; Rajchl et al., 2016; Hsu et al., 2019). By alerting users to precise spatial information about curbs, visual-based curb segmentation can augment safe navigation, affording greater independence to PBLV as they traverse diverse landscapes. This makes visual-based curb segmentation a fundamental task for assisting navigation for PBLV.

The advent of foundation models marks a new epoch in machine learning (Zohar et al., 2023). These models are distinguished by training completed on extensive datasets, which imbues them with an unprecedented level of understanding and adaptability (Yuan et al., 2021). This comprehensive training allows foundation models to excel in various tasks from detecting objects in complex scenes to accurately segmenting specific elements from an image (Yu et al., 2023). They can be applied across diverse domains without the need for additional task-specific training. This attribute significantly reduces the time and resources required for development and application.

This paper aims to explore the application of foundation models in the specific task of curb segmentation. Foundation models, which have shown robust performance across a wide array of tasks (Zhang et al., 2024; Shah et al., 2023; Bommasani et al., 2021; Wu et al., 2023; Salin et al., 2023), present a promising approach to this challenge, as well as many other specific areas for PBLV. However, currently, there is no systematic comparison of foundation model performance on curbs. Thus, the following question emerges: is curb detection and segmentation challenging for foundation models? Where do models struggle, and how might this be addressed?

To answer these questions, we introduce a new dataset for benchmarking. The dataset includes 2133 photographs capturing curbs in a metropolitan area from a pedestrian perspective at different distances, orientations, and states of repair. This dataset also includes images with curb-like objects. The study team manually annotated all curb instances in each image frame as polygons.

The contributions of this paper are as follows:

1. We present a curb segmentation dataset with both positive images featuring curbs and negative images featuring curb-like objects. All curb instances are manually annotated as complex polygons instead of bounding boxes for more accurate model training and evaluation.

2. We benchmark the performance of several foundation models with a focus on their accuracy and inference time for performing curb segmentation. This is the first manuscript to systematically demonstrate the weaknesses and specific challenges of curb segmentation.

3. We demonstrate how specific changes to how the foundation models select areas for segmentation may improve their accuracy and explore these opportunities as potential solutions to help address the challenges encountered.

**Method**

*Curb Segmentation Dataset*

There are no existing large, high-quality datasets specially designed for curb segmentation. Previous datasets, such as those on Roboflow, are relatively small, with less than 500 images, and are inconsistent in annotation quality. These facts also apply to larger datasets such as BDD100K and ADE20K which include curb annotations (Yu et al., 2020; Zhou et al., 2019). The 2018 Berkeley Deep Drive (BDD) dataset consists of 70K training images, 10K validation images, and 20K testing images, under diverse weather conditions and at different daytimes in real-traffic urban streets (Zhou et al., 2020). However, BDD100K is challenging to utilize due to only showing the perspective of drivers and not pedestrians. ADE20K is a dataset for scene parsing which involves labelling each pixel in an image with segmentation masks. However, there are only 331 annotated curb instances in the ADE20k dataset. By comparison our dataset includes 2133 images and more than 2000 annotated curb instances, providing more robust information for training and evaluation of AI models.

Our dataset is divided into two parts: positive images and negative images. Curbs only appear in positive images and not in negative images. This enables us to test whether the foundation models could distinguish curbs from curb-like objects. There are 1620 positive images and 513 negative images.

Positive images were collected by two lab members around our health system in lower Manhattan, New York. They held a Nikon D500 20.9 Megapixel Digital Camera at chest level and approached curbs at different orientation angles to create different vantage points, which is shown in Fig. 1. A specific focus on curbs of various states of repair (poor/well maintained) were maintained across the acquisition period.

After positive images were collected, LabelMe was used to manually annotate curb instances in each image (Russell et al., 2008). LabelMe is a graphical image annotation tool for polygonal annotation of objects. We label curbs as polygons instead of bounding boxes to capture their shapes more accurately, as bounding boxes will not tightly fit the shape of many curbs, and curbs are often not rectangular in visual images, as shown in Fig. 2(b). Commonly, curbs on both sides of the road are captured in a photo, and we labelled them as shown in Fig. 2(c).

Negative images are selected from the NYU-VPR dataset, which includes images recorded in the same area as the positive images and contain objects easily misidentified as curbs (Sheng et al., 2021). As shown in Fig. 3, those objects include sidewalks, windowsills, cafe barriers, etc.

*Foundation models*

The field of object detection and segmentation has been transformed by foundation models like Contrastive Language-Image Pretraining (CLIP), Segment Anything, and Large Language and Vision Assistant (LLaVA) (Kirillov et al., 2023; Liu et al., 2023; Radford et al., 2021). These models represent an advancement of deep learning capabilities. CLIP, for instance, uniquely combines visual data with natural language processing, offering a more contextual and interpretative approach to object recognition.

Segment Anything stands out for its precision in delineating individual elements within images, a task that is traditionally challenging due to overlapping objects and varying backgrounds (Kirillov et al., 2023). LLaVA is an end-to-end trained large multimodal model that connects a vision encoder and LLM for general-purpose visual and language understanding (Liu et al., 2023). The deployment of these foundation models makes significant contributions to areas such as automated image analysis, enhancing the accuracy of surveillance systems, and refining user experience in digital platforms (Simmons & Vasa, 2023; Zhao et al., 2024).

Open vocabulary semantic segmentation involves using models that can segment and understand images based on a wide, flexible range of categories. These models are usually foundation models that have been fine-tuned for segmentation tasks, e.g., LISA, Lang-seg, and OVSeg (Lai et al., 2023; Medeiros, 2023; Liang et al., 2023). LISA is a model that builds upon the LLaVA model's language generation capabilities (Liu et al., 2023; Lai et al., 2023). It introduces a novel "SEG" token and employs an "embedding-as-mask" paradigm, which is key to its segmentation proficiency. Lang-seg harnesses the power of instance segmentation coupled with text prompts (Medeiros, 2023). Built on Segment Anything and the GroundingDINO detection model, Lang-seg stands out for its simplicity and effectiveness in object detection and image segmentation (Kirillov et al., 2023; Liu et al., 2023). OVSeg presents a novel approach by fine-tuning CLIP with masked image regions and their corresponding text descriptions (Liang et al., 2023). This model innovates by mining existing image-caption datasets using CLIP to align masked image regions with nouns in captions (Radford et al., 2021; Liang et al., 2023).

*Testing methods*

We selected three state-of-the-art foundation model based methods for the evaluation of curb segmentation performance on our dataset: LISA, Lang-seg, and OVSEG (Lai et al., 2023; Medeiros, 2023; Liang et al., 2023). We also explore improving the bounding box selection of Lang-seg to further increase performance. We name the improved method as "Modified-lang-seg" with details described below.

*Modified-lang-seg*

Lang-seg uses GroundingDINO to generate multiple bounding boxes along with confidence scores for various curbs within an image. Then, for areas within each bounding box, the Segment Anything model generates a mask segmenting the largest coherent object that spans the "curb" bounding box, which usually corresponds to the actual curb. However, as illustrated below, like other foundation models such as LLaVA and CLIP, Segment Anything does a poor job of distinguishing between curbs and sidewalks. As a result, it is crucial to select the most accurate bounding box to constrain Segment Anything's processing. We found that typically only one bounding box among all overlapping bounding boxes is the true positive - meaning it successfully contains the curb between the road and sidewalk. All other bounding boxes contain false positives, such as parts of the sidewalk or the road. Therefore, we introduced the following logic for curb bounding box selection: for any overlapping bounding boxes in an image, only keep the one with the highest confidence score, in essence filtering the results. This technique helps convert many false positive cases to true positives, as shown in Fig. 4.

**Benchmark Experiments**

*Evaluation*

We use multiple metrics to evaluate the performance of foundation models. The first metric is Intersection over Union. It measures the overlap between the ground truth and the predicted segmentation masks. Intersection is the area of overlap between the predicted segmentation and the ground truth. Union is the total area covered by both the predicted segmentation and the ground truth. We can calculate IoU using the formula below:

$$IoU = \frac{Area\ of\ Overlap}{Area\ of\ Union} = \frac{Intersection}{Intersection + Union - Intersection}$$

The second metric is the confusion matrix. It is a table used to describe the performance of a classification model on a set of test data for which the true values are known. Because our task is segmentation rather than classification, we slightly modify the definition of each item in the matrix. True positives and true negatives are positive and negative images that are correctly identified. False positive images are images where the model incorrectly segments non-curb areas as curbs. False negative images are images where the model does not segment curb areas as curbs. Here are the formulas for other metrics in the confusion matrix:

$$Accuracy = \frac{True\ Positive + True\ Negative}{Total\ Observations}$$

$$Precision = \frac{True\ Positives}{True\ Positives + False\ Positives}$$

$$Sensitivity = \frac{True\ Positives}{True\ Positives + False\ Negatives}$$

$$Specificity = \frac{True\ Negatives}{True\ Negatives + False\ Positives}$$

$$F1\ Score = 2 \times \frac{Precision \times Sensitivity}{Precision + Sensitivity}$$

The last metric is the inference time. For users with visual impairments, detecting changes in the location, shape, size, and distance of curbs in real-time allows users to be fully informed of their spatial relationship to the curb and how it updates as they dynamically move.

*Results*

Fig. 6, Fig. 7 and Table I show our main results focused on the performance and inference time of foundation models on our curb dataset.

**Performance**: Fig. 6 shows the performance of foundation models under the metric of average intersection over union (IoU). Modified-lang-seg leads with an IoU of 40.3%. Lang-seg, which does not feature our additional bounding box selection logic, has a lower IoU of 36.0%. LISA shows a further drop in IoU at 26.6%. OVSEG, however, lags considerably behind the other models with an IoU of only 1.39%. The low IoU of OVSEG appears to be attributable to its general inability to distinguish among curbs, sidewalks, and roads. This can be seen in confusion matrix I, where almost all segmentation results of OVSEG are false positives. Modified-lang-seg being the best performing of the four models appeared to be primarily driven by large increases (1.9x over lang-seg) in true positive curb segmentations, and large decreases (4x over lang-seg) in false positives (i.e., incorrectly segmenting non-curbs as "curbs").

Despite the performance differences between foundation models, their results all contain many false positives, as shown in Table I and illustrated in Fig. 8. Modified-lang-seg has the highest accuracy (0.5874) and precision (0.7035), demonstrating its superior performance by reducing false positives (fewest among the models). It also has a moderate performance in terms of sensitivity to segmenting actual curbs (0.6182) and crucially, the highest specificity of all models tested for *only* segmenting the actual curbs (0.5381). However, modified-lang-seg is prone to identify false negatives (highest among models). By comparison, Lang-seg has a much lower accuracy (0.3180) and precision (0.2331). It also has a very high sensitivity (0.9647), but very low specificity (0.1435), indicating while lang-seg's segmentations almost always contain the curb, they also contain a lot of non-curb areas (i.e., false positives). LISA has a similar accuracy (0.3666) and precision (0.2926). It has a relatively high sensitivity (0.6581) but its specificity remains low (0.2245). OVSEG has an extremely low accuracy (0.0530) and precision (0.0079). It has perfect sensitivity on paper (1.0000), meaning it identified all positives and did not miss any curbs (0 false negatives), but its specificity (0.0458) is extremely low, suggesting it almost always incorrectly predicts negative areas as curbs.

**Inference Time**: The average inference times for these foundation models was fastest for OVSEG at 3.70 seconds, followed by lang-seg at 4.54 seconds and Modified-lang-seg at 4.77 seconds, while LISA more than doubled this time at 10.85 seconds.

**DISCUSSION**

Following our experiments, we found that curbs are a difficult spatial hazard to accurately identify using state-of-the-art foundation models. More specifically, curb segmentation is challenging for all tested models due to their poor performance distinguishing between curbs, non-curb areas, and curb-like objects. In our results, there are many false positives found with all foundation models, however, modifications made to the modified-lang-seg were found to substantially reduce these issues. It was also found that the fastest of these foundation models takes 3.7 seconds on average to segment an image, which poses a significant barrier for applications designed to assist PBLV during dynamic tasks. In such applications, real-time feedback is crucial for safe navigation and immediate decision-making in dynamic environments, and hence, achieving this task with faster inference times will be essential to increasing their functional utility. Our results provide insights into both current issues and potential solutions for effectively leveraging current advances in AI to facilitate tasks crucial for safe navigation for PBLV.

*Weaknesses of Unmodified Foundation Models*

In our comparison of foundation models for curb segmentation, we found comparable performance in terms of their accuracy and precision across two unmodified approaches – lang-seg and LISA, with a substantial drop-off in performance for OVSEG. For images containing curbs, Lang-seg and OVSEG have difficulties in distinguishing between sidewalks and curbs (see Fig. 8), while LISA does not have this problem as significantly, it is instead more resistant to segmenting actual curbs, resulting in the higher false-negative scores. Lang-seg does well in not misattributing the road as part of the curb segmentation. For negative images, these models have difficulties distinguishing between windowsills, ground surfaces, cafe barriers, and curbs. Overall, foundation models struggle to accurately discern between curbs, sidewalks, and objects that resemble curbs.

While Lang-seg and LISA are comparable in terms of curb segmentation accuracy, they differ substantially in terms of inference time. The extended inference time for LISA (10.85 seconds) could be attributed to its more complex reasoning segmentation process. Although the prompt for all three models is the same, LISA may still require additional time for prompt analysis. For performance on a repetitive task, this drawback could substantially undermine its use as a practical tool for real-time feedback for PBLV. Ultimately, real-time feedback for PBLV with foundation models will require improvements to inference time, either through direct improvements to the models, or by using additional computational resources (e.g., high-performance servers) for these services. It is untenable for an end user to implement feedback offered from an assistive technology that is presented outside of physiologic windows or temporal epochs that are too protracted and misaligned with biomimetic principles.

*Adapting Foundation Models for Assistive Technologies*

In the current context, low values on either sensitivity or specificity in the lang-seg, LISA, and OVSEG models would render them unreliable as assistive technologies for PBLV. For example, if the model performs poorly in terms of specificity, the user will constantly be informed about false positives "curbs", struggling to distinguish true negatives. These errors are not just minor oversights but significant misclassifications that undermine the utility of the models in practical scenarios, such as autonomous navigation and pedestrian safety applications, in which precise object segmentation is crucial. The challenge lies in enhancing the models' ability to contextualize and uniquely identify curbs by understanding the subtle distinctions that differentiate them from other urban features.

Fortunately, with simple modifications to the bounding box selection for lang-seg, overall performance can be boosted substantially. By only selecting the highest confidence bounding box of multiple overlapping boxes for segmentation, there is an increase in true positives, true negatives, as well as a substantial decrease in false positives. This performance increase costs only an additional 230ms in terms of inference time. There is another downside however, as this modification also appears to increase false-negative scores. This indicates that our modification now makes lang-seg overly conservative in segmenting curbs. As such, further alterations to our specific modification may help balance these factors better.

Ultimately, we attribute Modified-lang-seg's superior performance across all tested models down to two main reasons. Firstly, Segment Anything's superior ability to segment areas within the bounding boxes that best correspond to actual curbs, which would improve the model's Intersection-over-Union scores. Secondly, given the filtering performed to the bounding box selection, results appear to reflect substantial reductions to false positive detections and subsequent erroneous segmentations. However, as discussed later, while this approach appears to be beneficial to foundation models, this process may also extend to other approaches as well.

*Integrating Models for Greater Efficiency (combinatorial AI)*

Future work to enhance assistive technologies for PBLV appears promising, particularly with the potential for advances as seen in Lang-seg. Lang-seg's integration of bounding boxes and SAM demonstrates a notable ability to accurately segment curbs, suggesting that this approach could be pivotal in developing local systems for real time curb segmentation. Recently, smartphone-compatible versions of YOLOv8 provide object detection (but not yet segmentation) that run in real-time with YOLOv8-nano running at 15ms and YOLOv8-XL running at 51ms inference time on an iPhone 14 pro hardware (Gwak, 2019). In addition, methods such as MobileSAM (Zhang et al., 2023) show substantial improvements to the efficiency of Segment Anything Model and can reach 12ms inference time per image (Zhang et al., 2023). The combination of highly trainable smartphone-compatible object detection methods providing bounding boxes for rapid segmentation models could be an alternative path to providing real-time curb segmentation. In combination with the introduction of our curb segmentation dataset, this could help make curb segmentation more accessible using alternative methods.

**CONCLUSION**

In conclusion, this study highlights significant challenges in a primary spatial hazard for persons with disability, particularly persons with blindness and low vision. More specifically, curb segmentation with foundation models was found to be quite poor, revealing their limitations in differentiating between curbs and similar streetscape features. Despite these challenges, the adaptation of a modified-lang-seg model was demonstrated to be effective in reducing false positives, showcasing the potential for tailored approaches in enhancing model performance. Presently, processing speeds remain a barrier, underscoring the need for advances that can deliver real-time feedback essential for safe navigation of persons who are blind or low vision. This work not only improves our understanding of the technological hurdles but also sets a clear direction for future enhancements to increase the utility and effectiveness of assistive technologies focused on safe navigation.

## Acknowledgements

We would like to thank Eliot Sperling for his assistance in acquiring superb photos of NYC curbs.

|                | Modified-lang-seg | Lang-seg | LISA   | OVSEG  |
|----------------|-------------------|----------|--------|--------|
| True Positive  | 847               | 437      | 460    | 16     |
| True Negative  | 406               | 241      | 322    | 97     |
| False Positive | 357               | 1439     | 1112   | 2020   |
| False Negative | 523               | 16       | 239    | 0      |
| Accuracy       | 0.5874            | 0.3180   | 0.3666 | 0.0530 |
| Precision      | 0.7035            | 0.2331   | 0.2926 | 0.0079 |
| Sensitivity    | 0.6182            | 0.9647   | 0.6581 | 1.0000 |
| Specificity    | 0.5321            | 0.1435   | 0.2245 | 0.0458 |
| F1 Score       | 0.6581            | 0.3754   | 0.4051 | 0.0156 |

**Table 1.** Confusion Matrices with Sensitivity and Specificity

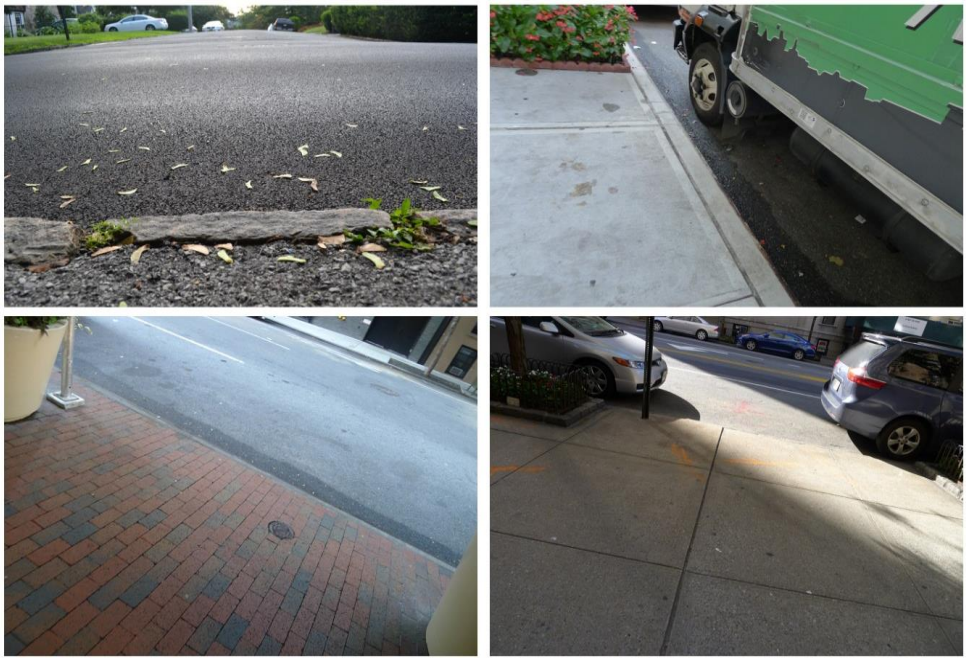

**Figure 1.** Curbs are captured at different angles in positive images.

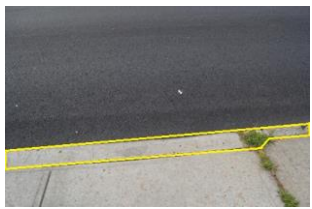 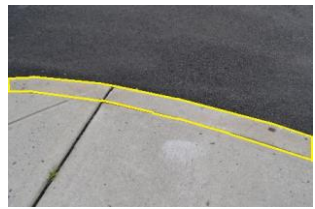

(a) Rectangular curb.   (b) Polygonal curb.

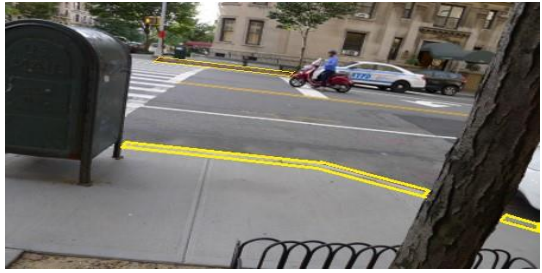

(c) Multiple polygonal curbs

**Figure 2.** Example images with annotated curb instances.

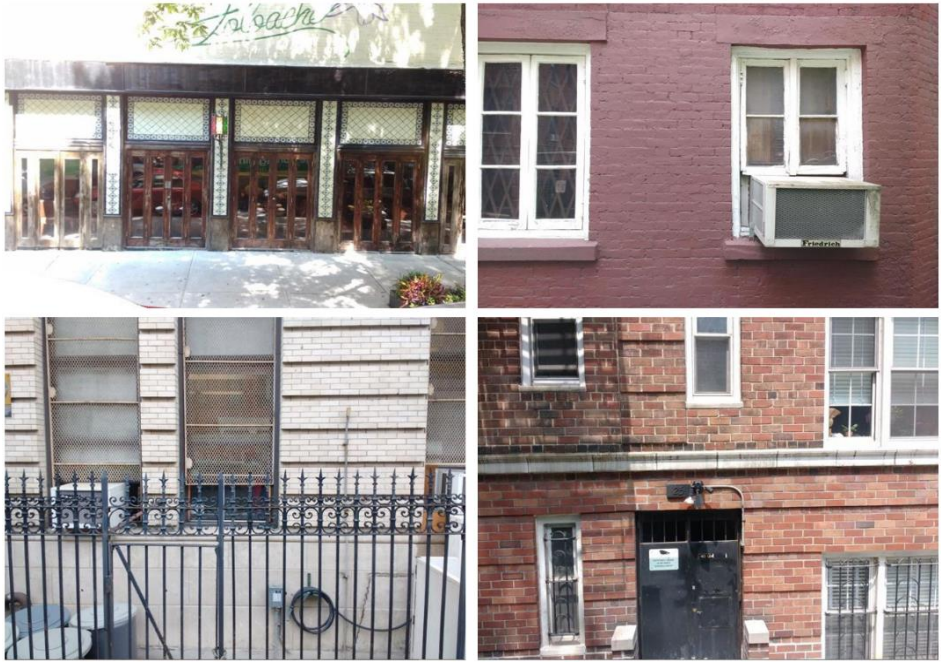

**Figure 3.** Example negative images include curb-like objects such as sidewalks and windowsills.

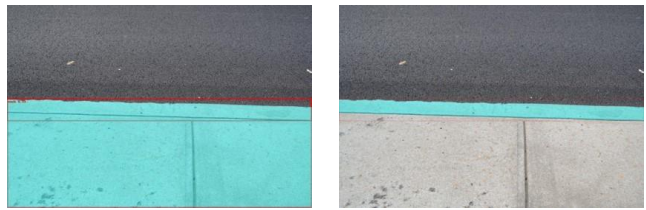

**Figure 4.** Improved result of Modified-lang-seg model (right compared to left).

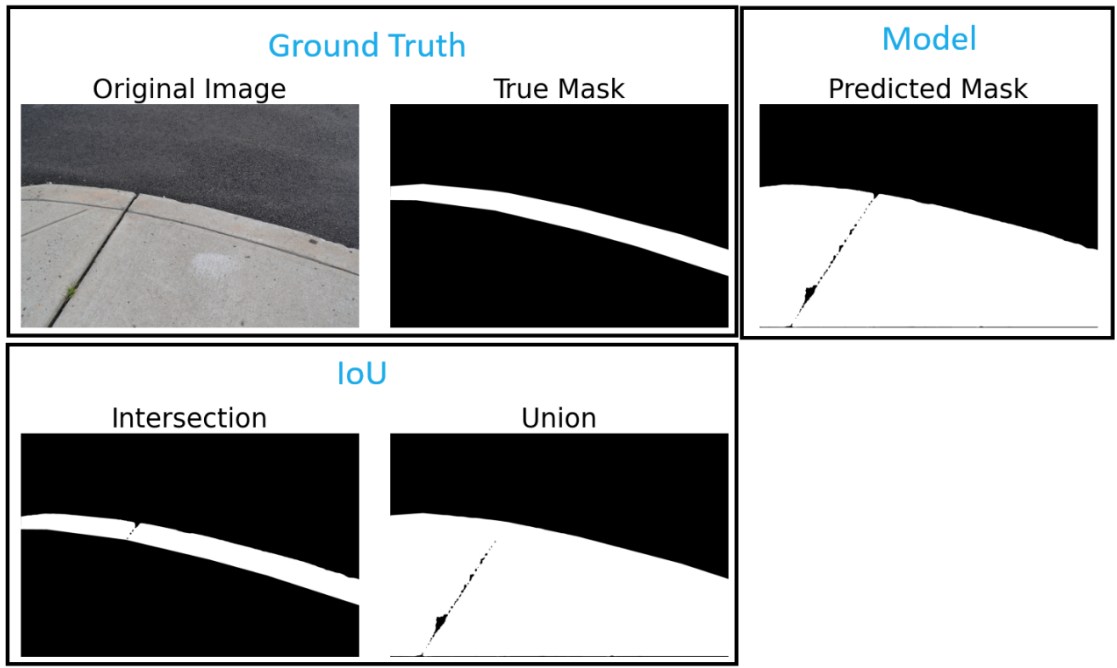

**Figure 5.** An example demonstrating Intersection over Union.

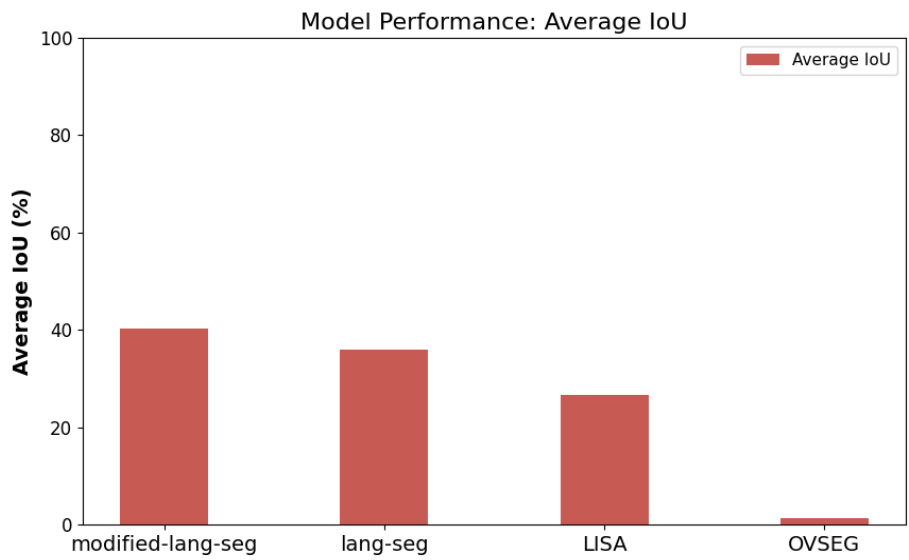

**Figure 6.** Average Intersection over Union of each foundation model.

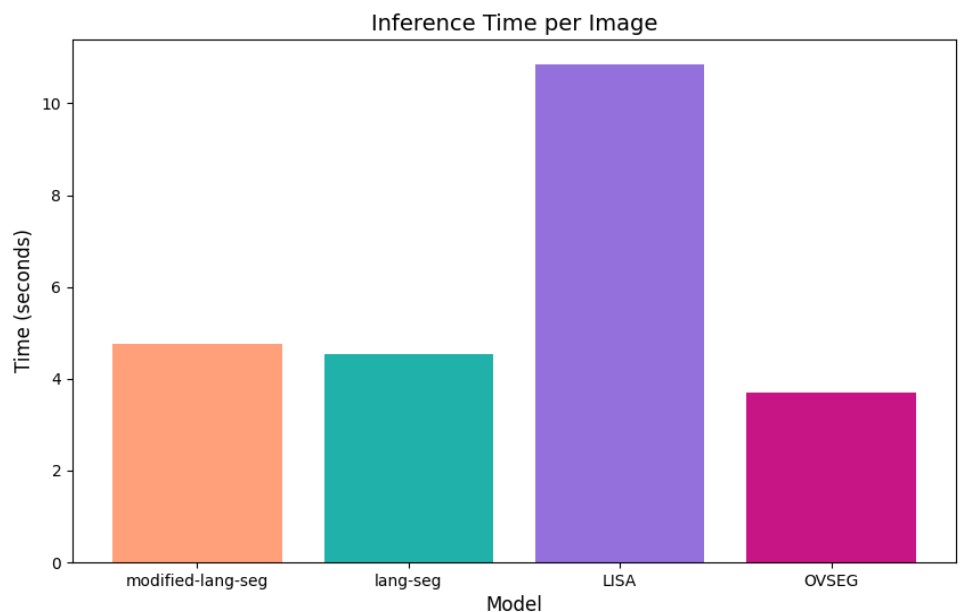

**Figure 7.** Average inference time of each foundation model.

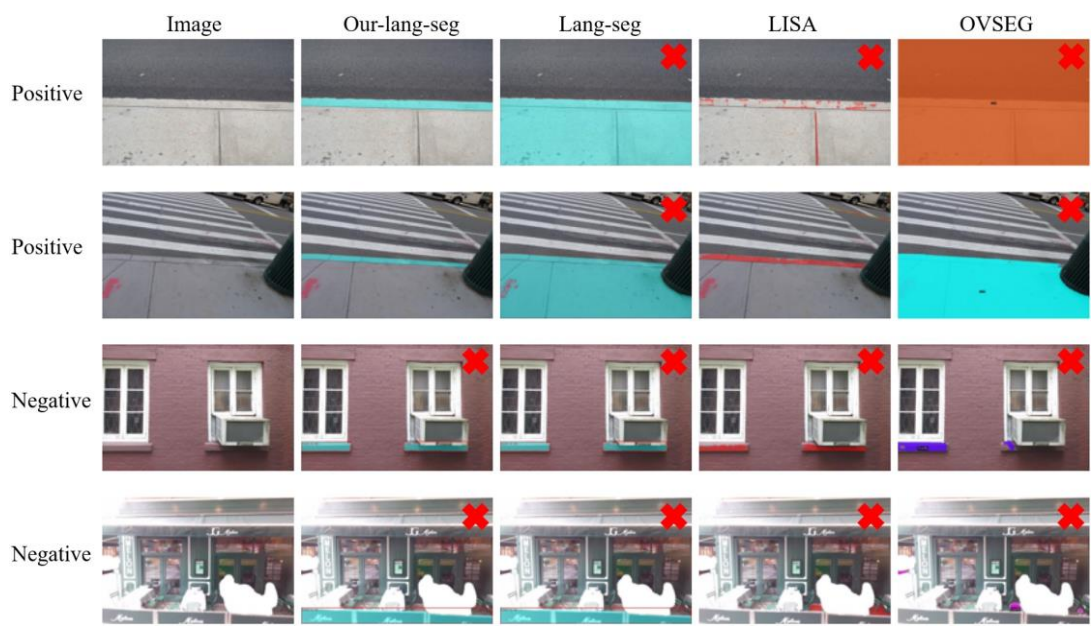

**Figure 8.** We visualize object segmentation results for Modified-lang-seg, Lang-seg, LISA, and OVSEG. We randomly picked two positive images and two negative images. The red cross means the result is false positive or false negative.